\documentclass[letterpaper, 10 pt, conference]{ieeeconf}  %
\IEEEoverridecommandlockouts                             
\pdfoutput=1
                                                         
\usepackage[nolist]{acronym}
\usepackage[usenames,dvipsnames,svgnames,table]{xcolor}
\usepackage{soul}
\usepackage{mathtools}
\usepackage{verbatim}
\usepackage[pdftex]{graphicx}
\usepackage{epsfig,subcaption}
\usepackage{amsmath,amssymb,amsfonts,bm,amscd} 
\usepackage{mathrsfs}
\usepackage{setspace}
\usepackage{multirow}
\usepackage{fancyhdr}
\usepackage{algorithm} 
\usepackage{psfrag}
\usepackage{multirow, makecell}
\usepackage{svg}
\usepackage{cite}
\usepackage{hyperref}
\usepackage{color}
\usepackage{cleveref}
\usepackage{dblfloatfix}
\usepackage[export]{adjustbox}

\usepackage{xcolor}
\usepackage{framed}
\usepackage{lipsum}

\colorlet{shadecolor}{yellow!50}

\usepackage{algorithm}
\usepackage{algpseudocode}

\newcolumntype{C}[1]{>{\centering\let\newline\\\arraybackslash\hspace{0pt}}m{#1}}
\usepackage{xcolor,colortbl}
\definecolor{lightgreen}{RGB}{173,213,176}
\definecolor{lightyellow}{RGB}{255,255,204}
\definecolor{lightblue}{RGB}{204,255,255}
\definecolor{lightred}{RGB}{255,204,204}

\newcommand{\xxnote}[3]{}
\ifx\hidenotes\undefined
  \usepackage{color}
  \renewcommand{\xxnote}[3]{\color{#2}{#1: #3}}
\fi

\RequirePackage{tikz} 

\DeclareMathOperator*{\minimize}{{min}}

\begin{acronym}
    \acro{NMPC}{nonlinear model predictive control}
    \acro{MPC}{model predictive control}
    \acro{VT-NMPC}{visual tracking nonlinear model predictive control}
    \acro{PAMPC}{perception-aware model predictive control}
    \acro{ROI}{region of interest}
    \acro{FOV}{field of view}
    \acro{VS}{visual servoing}
    \acro{PBVS}{position-based visual servoing}
    \acro{PBVS}{position-based visual servoing}
    \acro{IBVS}{image-based visual servoing}
    \acro{STL}{standard triangle language}
    \acro{KKT}{karush–kuhn–tucker}
    \acro{TSP}{traveling salesman problem}
\end{acronym}

\title{\LARGE \bf
Visual Tracking Nonlinear Model Predictive Control Method for Autonomous Wind Turbine Inspection
}

\author{ 
Abdelhakim Amer, Mohit Mehndiratta, Jonas le Fevre Sejersen, Huy Xuan Pham, Erdal Kayacan
\thanks{A. Amer, J. le Fevre and H. X. Pham are with Artificial Intelligence in Robotics Laboratory (AiR Lab), the Department of Electrical and Computer Engineering, Aarhus University, 8000 Aarhus C, Denmark, \{abdelhakim, jonas.le.fevre, huy.pham\} @ece.au.dk. Mohit Mehndiratta is with GIM Robotics, Espoo, Finland, mohit.mehndiratta@gimrobotics.fi. E. Kayacan is with the Automatic Control Group, Department of Electrical Engineering and Information Technology, Paderborn University, Paderborn, Germany,  \{erdal.kayacan\} at uni-paderborn.de.}}

\begin{document}
\maketitle
\begin{abstract}

Automated visual inspection of on-and offshore wind turbines using aerial robots provides several benefits, namely, a safe working environment by circumventing the need for workers to be suspended high above the ground, reduced inspection time, preventive maintenance, and access to hard-to-reach areas. A novel nonlinear model predictive control (NMPC) framework alongside a global wind turbine path planner is proposed to achieve distance-optimal coverage for wind turbine inspection. Unlike traditional MPC formulations, visual tracking NMPC (VT-NMPC) is designed to track an inspection surface, instead of a position and heading trajectory, thereby circumventing the need to provide an accurate predefined trajectory for the drone. An additional capability of the proposed VT-NMPC method is that by incorporating inspection requirements as visual tracking costs to minimize, it naturally achieves the inspection task successfully while respecting the physical constraints of the drone. 
Multiple simulation runs and real-world tests demonstrate the efficiency and efficacy of the proposed automated inspection framework, which outperforms the traditional MPC designs, by providing full coverage of the target wind turbine blades as well as its robustness to changing wind conditions. The implementation codes\footnote{\url{https://www.github.com/open-airlab/VTNMPC-Autonomous-Wind-Turbine-Inspection}} are open-sourced.

\end{abstract}

\section{INTRODUCTION}


Wind turbine blades operate in harsh working conditions, underlying exposure to high centrifugal loads, erosion, lightning strikes, and bird hits. Thus, their condition needs consistent monitoring via regular inspections. Traditionally, human operators conduct arduous visual inspections, which are expensive and underlie significant risks, as the inspectors are suspended high above the ground. Consequently, piloted drones are deployed due to lower operating costs and reduced downtime of turbines by performing faster inspection \cite{NLR}. However, manual inspections heavily rely on pilots' experience and skills. Hence, the need for an automated inspection framework that substitues a skilled human pilot is evident.

There are several challenges in the automated drone-based inspection of wind turbine blades. One of them is maintaining a specific distance from the blades’ surface. A large distance negatively affects image resolution, while being too close involves the risk of crashing into the blade, especiallty when exposed to strong winds.
Another challenge is that the drone should be perpendicular to the blades’ surface for an optimal view angle of the region of interest. This challenge underlies two aspects: (i) obtaining the relative viewing angle to the surface gets tricky for a trajectory with desired position and heading attributes, and (ii) maintaining the desired relative view requires consistent rejection of operational disturbances from the controller.

The core of the proposed methodology is the underlying surface trajectory -- comprising of an optimal sequence of inspection surfaces -- that replaces an explicit position and heading trajectory for the drone. Such a novel approach essentially circumvents closely tracking a predefined trajectory that is no longer appropriate to achieve full inspection coverage due to operational disturbances. To start with, a global distance-optimal sequence of inspection surfaces is obtained via a greedy search-based optimization strategy over each wind turbine blade surface. Next, a novel \ac{VT-NMPC} method is proposed which manipulates the drone's pose to precisely cover each inspection surface. Owing to the well-crafted objective function, the \ac{VT-NMPC} method allows the drone to consistently correct its relative pose to the blade's surface, thus maintaining the desired distance and optimal viewing angle to the inspected turbine blade. 
The contributions of this work are as follows:

\begin{itemize}
    \item 
    An automated wind turbine inspection framework that renders end-to-end inspection by taking turbines' global position and dimensions as inputs.
 

    \item A novel \ac{NMPC} method with visual tracking objectives, rendering an optimal relative drone's pose to the inspection surface at all times. 
    
    \item Implementation and analysis of the proposed method
on a customized drone to demonstrate its real-world applicability.

\end{itemize}

\begin{figure*}
    \centering
    \includegraphics[width=1\textwidth]{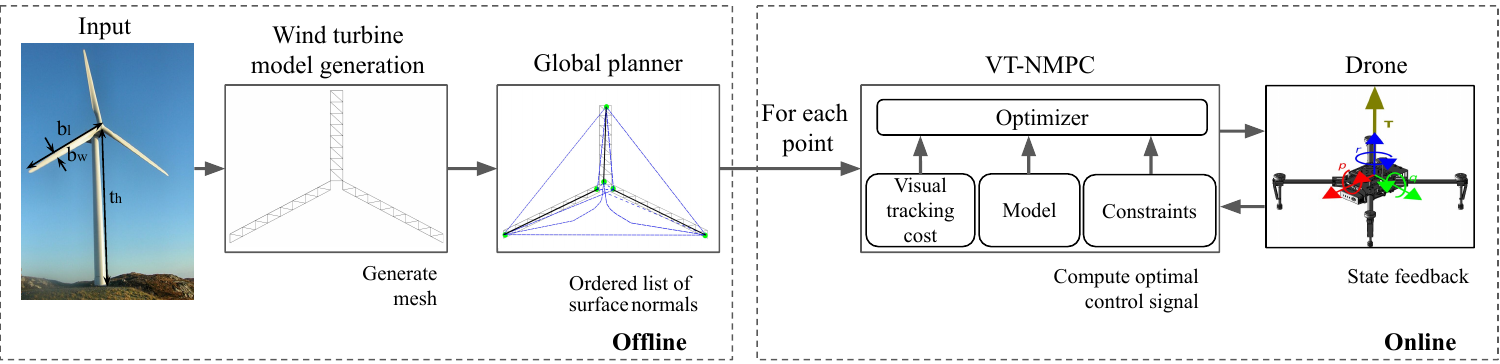}
    \caption{Overview of the automated inspection framework modules. The user specifies the dimensions and location of the turbines and the output is the motor commands for the drone. The model generation and global planning are calculated pre-flight (offline), while the optimization and control are performed onboard (online).} 
    \label{fig:abstract}
\end{figure*}

The rest of this paper is organized as follows: Section \ref{sec:RelatedWork} introduces the state-of-the-art inspection methods. Section \ref{sec:methodology} presents the problem formulation and illustrates the proposed automated inspection framework. The mathematical model of the inspection drone is presented in Section \ref{sec:robot} followed by the \ac{VT-NMPC} problem formulation in Section \ref{sec:design}. The simulation results are presented in Section \ref{sec:results}, followed by real-world implementation experiments in Section \ref{sec:results_r}. Finally, some conclusions are drawn from this work in Section \ref{sec:conclusion}.

\section{Related Work}
\label{sec:RelatedWork}

Several approaches tried to address the problems of state estimation, planning, and control for aerial vehicles in challenging outdoor scenarios \cite{jakob,micha,aruco}. For autonomous wind turbine inspection, trajectory-tracking-based approaches are proposed in the literature wherein an optimal waypoint sequence comprising of 3D position and heading, is generated beforehand \cite{CPP,3dmodeling,SIP}. Howbeit, disturbances such as wind render the predefined trajectory insufficient to achieve full coverage of the inspection area.
Alternatively, other solutions, \cite{mohit,lidar} generate the inspection trajectory on the fly by utilizing relative distance measurements. The main limitation of these methods is the global suboptimality of the generated trajectory. Additionally, distance-optimality is a desired feature for an automated wind turbine inspection framework because of the limited battery life of the drones.

Other approaches, such as  \Ac{VS} methods provide a trajectory that allows the drone to maintain a collision-free sight to a given point of interest. For instance, in \cite{target_aware}, an optimal trajectory for a quadrotor is obtained by solving a nonlinear optimization problem, followed by the trajectory tracking via \ac{NMPC}. Another method that provides an optimal trajectory by developing a path parameterization algorithm for quadrotors, considering their limited \ac{FOV}, is proposed in \cite{mit}. 
As such, all these \ac{VS} methods treat the trajectory generation and tracking as separate tasks. Conversely, a more efficient way is to generate the trajectory as part of the control problem. This approach is illustrated in \cite{VS_MPC}, 
wherein they design a linear MPC that renders trajectory tracking while generating {\ac{VS}}-based velocity profiles to navigate the quadrotor.
Another work based on a similar principle is illustrated in \cite{falanga2018pampc}, where the \ac{PAMPC} method is proposed. This method adds a cost to the problem formulation of \ac{NMPC} to maintain a target within the \ac{FOV}-center while minimizing image blur. 

The above methods plan a local path that fulfills a predefined visual goal rather than resulting in a global inspection trajectory. Therefore, they are insufficient on their own to realize all inspection-related goals, such as finding a global distance-optimal path and achieving total coverage of the inspection surface. In the next section we present our hierarchical inspection framework, wherein an offline distance-optimal planner results in a global trajectory, followed by the VT-NMPC method rendering local planning. 
\section{Automated Inspection Framework}
\label{sec:methodology}
The turbine blades need to be inspected in two phases. In the first phase, the wind turbine rests in an inverted Y configuration, wherein the top blade is oriented vertically upward, while the other two are at a $120$ degree angle from the vertical. In order to inspect the surfaces pointing towards the ground, the wind turbine is rotated $120$ degrees, and a re-planning is performed on the last two surfaces. The inspection framework is demonstrated for a Vestas V100, having tower height ($t_h$) of $120$m, blade length ($b_l$) of $50$m, and blade width ($b_w$) of $3$m (See Fig. \ref{fig:abstract}).
An overview of the automated inspection framework is presented in Fig. \ref{fig:abstract}.

\subsection{Triangular Mesh Generation}

First, a simplified triangular mesh representation of the turbine blades is created. For this purpose,  wind turbine dimensions, $b_l$, $b_w$, and $t_h$, are given as inputs. Three cuboids of dimensions $b_l$ $\times$ $b_w$ $\times$ $\frac{b_w}{3}$ are then created with a $120^{\circ}$ relative angle between them. The cuboids represent the turbine blades in a simplified form, wherein each cuboid comprises right-angle triangular elements on each surface.  This simplified model overestimates the blade's width as it does not account for the tapering towards the tip. However, introducing this simplification facilitates generating a model of any commercial wind turbine, irrespective of design or size, thus rendering the proposed method generic. It should be noted that the mesh generation process can be customized to support any desired initial wind turbine configuration, and is not limited to the Y configuration. 

\subsection{Wind Turbine Inspection Path Planner}

\begin{figure}
    \centering
   
    \includegraphics[width=0.5\textwidth]{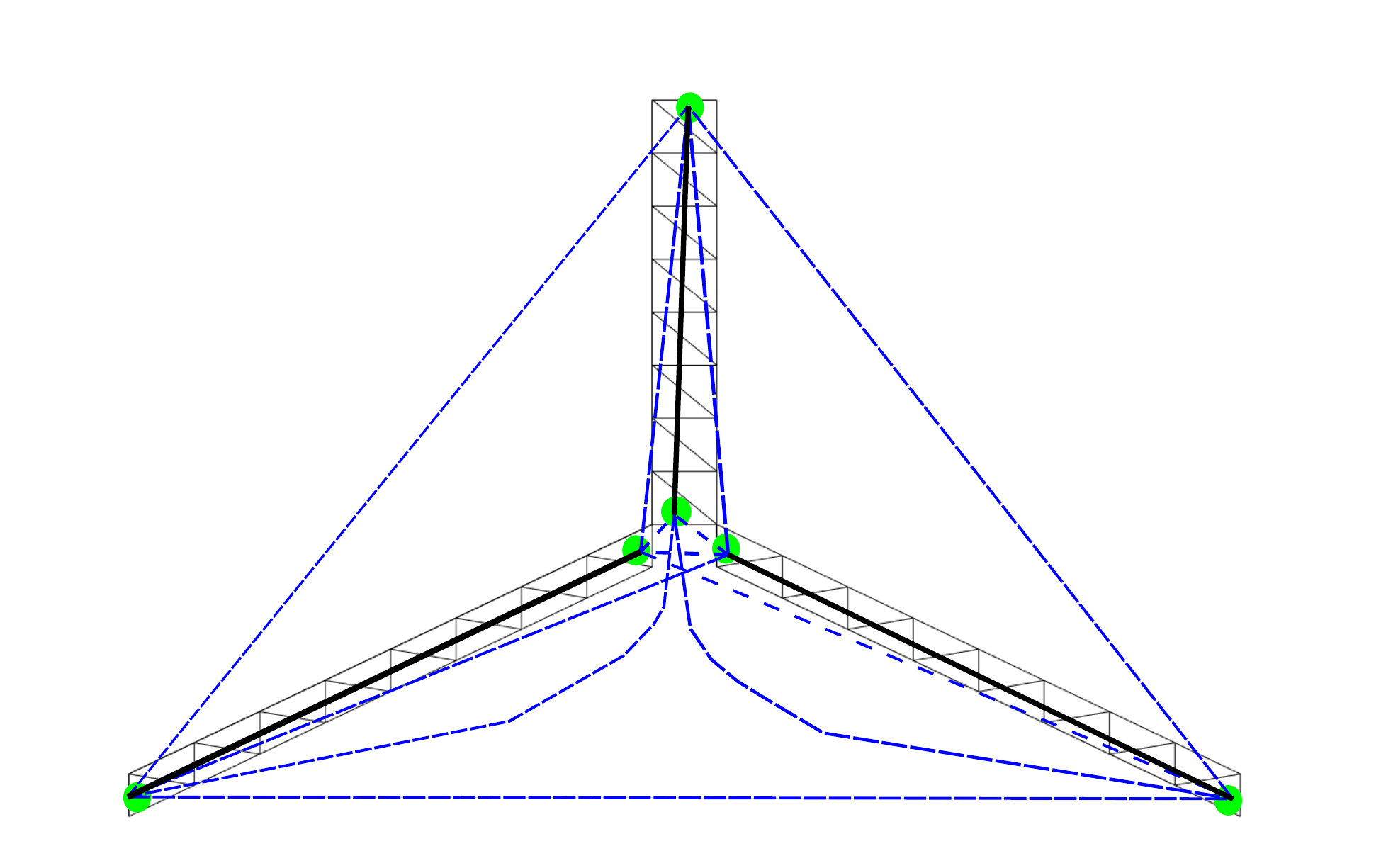}
    \caption{
   A topological representation of our graph-based search algorithm for distance-optimal path planner. A node (green) is placed at the end of each blade's surface, which are all connected by (blue) edges to form a graph. The goal is to find the sequence of nodes to visit that minimizes the total distance traveled while satisfying the imposed constraints. }
    
    \label{fig:planner}  
\end{figure}

In essence, the function of this planner is to obtain a distance-optimal sequence of inspection points, which are the centers of each triangular mesh surface, utilizing the generated simplified mesh model. For this purpose, the path-planning algorithm underlies three steps: Clustering, ordering, and interpolation. Within the clustering step, each triangular element from the wind turbine model is grouped into different surfaces based on its centroid location and value of the surface normal. The wind turbine model then consists of twelve clusters (surfaces) in total, four for each blade. Then, in the second step, two nodes are defined for each cluster or blade surface, whereby the nodes are located at each end of the surface (root and tip). Thus, a graph structure is created by connecting each node, as shown in Fig.~\ref{fig:planner}, where the edge length connecting the nodes is the cost to be minimized. Subsequently, the ordered list for the node is obtained via solving a \ac{TSP}, with a constraint that renders the entry and exit node to always belong to the same cluster. This constraint facilitates sequential surface inspection while ensuring that the drone does not jump between surfaces. Since the generated graph comprises a relatively small number of nodes ($24$), we adopt a brute force algorithm to solve the \ac{TSP}, thus guaranteeing distance optimality. Finally, the ordered list of intermediate points and corresponding normals between the nodes are found via linear line fitting and interpolation techniques. A very simplified version of the wind turbine-specific path planner algorithm is provided as pseudocode in Algorithm \ref{alg:1}.
Similar to \cite{SIP}, the proposed path planner uses a mesh generated model of the turbine for computing the distance-optimal path. However, this work exploits the wind turbine structure, wherein the graph is created based on surfaces rather than individual triangles. This simplification reduces the number of nodes by several orders of magnitude, thus significantly lowering the computation time. 
Yet, the simplification compromises the generality as a universal planner, restricting it to only wind turbine inspection.
While  \cite{SIP} is a general planner for any structure, the output of the planner is not compatible with the proposed controller. The output from \cite{SIP} is the reference point of the inspection drone, while the proposed planner and controller are working with reference of the inspection point on the mesh surface. One could project the drone's reference point onto the inspection mesh, but as \cite{SIP} computes optimal position by utilizing the FOV of the camera, there is no direct correlation between the projected drone references points and the inspection point. In conclusion, a new planning method was needed in order to fully utilize the potential of the proposed controller.

\begin{algorithm} 
\caption{Wind turbine specific path planner}
\label{alg:1}
\begin{algorithmic}
   \State \textbf{A) Create the graph (clustering)}
   \State $groups \leftarrow$ group elements by normals $\&$ centroid location
   \State $groups \leftarrow$ sort elements in each group by z-value
\State \textbf{B) Solve the TSP via brute force (ordering)}


    \While{there are more permutations of the tour, T }
        \State T = []
        \State shuffle groups
        \For{$group$ in $groups$}  
        \State		T append one of end nodes in $group$
		\State	T append other end node in $group$
		\EndFor
		\If{T is not new}
		\State  continue
		\EndIf
		\If {distance(T) $<$ $shortest\_distance$}
            \State $best\_tour \leftarrow$ T 
            \State $shortest\_distance \leftarrow$ distance(T)
        \EndIf
    \EndWhile

    \State \textbf{C) Find intermediate points (interpolation)}
    \State $intermediate\_points \leftarrow$ line fitting and interpolation 
    
\end{algorithmic}

\end{algorithm}

\subsection{Visual Tracking Nonlinear Model Predictive Control }



An NMPC is an optimization-based control method that is commonly preferred for robotic applications {\cite{mpc_survey, Mohit2017Receding, Mohitacc2018, Mehndiratta2019,quadplux,tilt,mohit_fault_tolerant_c,mohit_3dprint,tractor, kraus_mpc, arm_mpc}}. It calculates control actions utilizing constrained optimization and a model-based prediction. In this work, based on the given inspection point trajectory and the state measurements, the designed \ac{VT-NMPC} method calculates the desired attitude rates and thrust, which are given as inputs to the PID controller performing the actuator control. Moreover, the \ac{VT-NMPC} method facilitates visual coverage of the inspected structure by satisfying some visual tracking costs that are specified later in Section \ref{sec:design}.
Unlike other MPC controllers, the designed \ac{VT-NMPC} method does not require the desired position trajectory for the drone. Note that the proposed VT-NMPC is a generic inspection controller and is not limited to wind turbine inspection. The proposed method has two advantages. Firstly, the waypoints are not always available for an inspection problem, whereas it is straightforward to provide the inspection surface's position as input. Secondly, safety is ensured throughout the flight, as the \ac{VT-NMPC} method consistently optimizes the objective of keeping the drone at a specific distance from the inspection surface.




\section{Inspection drone}\label{sec:robot}
We present the dynamic model of the utilized quadrotor in Newton-Euler format \cite{deepmodel_mohit}, while replacing Euler angles with quaternion. The translational kinematics is obtained using the transformation from body frame ($\mathcal{F}_B$) to Earth-fixed frame ($\mathcal{F}_E$) as follows:
\begin{align} \label{eq:kin_tilt}
	& \left[ \begin{array}{c} \dot{x} \\ \dot{y} \\ \dot{z} \end{array} \right] = R_{EB} \left[ \begin{array}{c} u \\ v \\ w \end{array} \right],
\end{align}
where $x$, $y$, $z$ represent the translational position that is defined in frame $\mathcal{F}_E$, while $u$, $v$, $w$ are the translational velocities and are defined in the frame $\mathcal{F}_B$ and, finally, $R_{EB}$ represents the rotation matrix between frames $\mathcal{F}_E$ and $\mathcal{F}_B$, 
%
%
associated with the unit quaternion vector $\mathbf{q}_{EB}=q_x, q_y, q_z, q_w$. The rigid-body dynamic equations in the body-fixed coordinate are given as follows, assuming the quadrotor to be a point mass such that all the forces act at the center of gravity.
\begin{subequations} \label{eq:force}
\begin{align}
	\dot{u} &= r v - w q + 2 g (q_x  q_z - q_w  q_y), \label{eq:force_x} \\
	\dot{v} &= p w - r u - 2 g (q_y  q_z + q_w  q_x),  \label{eq:force_y} \\
	\dot{w} &= q u - p v - g (1 - 2 q_x  q_x - 2  q_y  q_y)  + \frac{1}{m} T, \label{eq:force_z} 
\end{align}
\end{subequations}
where $p$, $q$, $r$ are the angular body rates defined in the body frame, $T$ is the total thrust force generated by the drone's propellers in the body frame, and $m$ is the quadrotor's total mass. Additionally, the following represents the relation between the global angular rates in $\mathcal{F}_E$ and the body rates in $\mathcal{F}_B$:
\begin{subequations} \label{eq:rates}
\begin{align}
	\dot{q_x} &= 0.5 (p q_w + r q - q q_z), \\
    \dot{q_y} &= 0.5  (q q_w - r q_x + p q_z), \\
    \dot{q_z} &= 0.5 (r q_w + q q_x - p q_y), \\
    \dot{q_w} &= 0.5 (-p q_w - q q_y - r q_z).
\end{align}
\end{subequations}
%

Finally, we can rewrite the lumped nonlinear dynamic model of the quadrotor at a high level in its discretized form as follows:
\begin{align} \label{eq:model_eqs}
    \mathbf{x}_{k+1} = \mathtt{f_d}(\mathbf{x}_k,\mathbf{u}_k),  \quad
\end{align}
where $\mathbf{x} \in \mathbb{R}^{10}$ and $\mathbf{u} \in \mathbb{R}^{4}$, are the state and control vectors and they comprise of:
\begin{align}
    \mathbf{x} &= [x,y,z,u,v,w, q_x,q_y,q_z,q_w]^T, \quad  \\
    \mathbf{u} &= [p,q,r,T]^T.
\end{align}
The state function is denoted by $\mathtt{f_d}(\cdot,\cdot)$: $\mathbb{R}^{10} \times \mathbb{R}^{4} \rightarrow \mathbb{R}^{10}$.

\section{Proposed VT-NMPC Formulation}
\label{sec:design}

\begin{figure}
    \centering
    \includegraphics[width=0.5\textwidth]{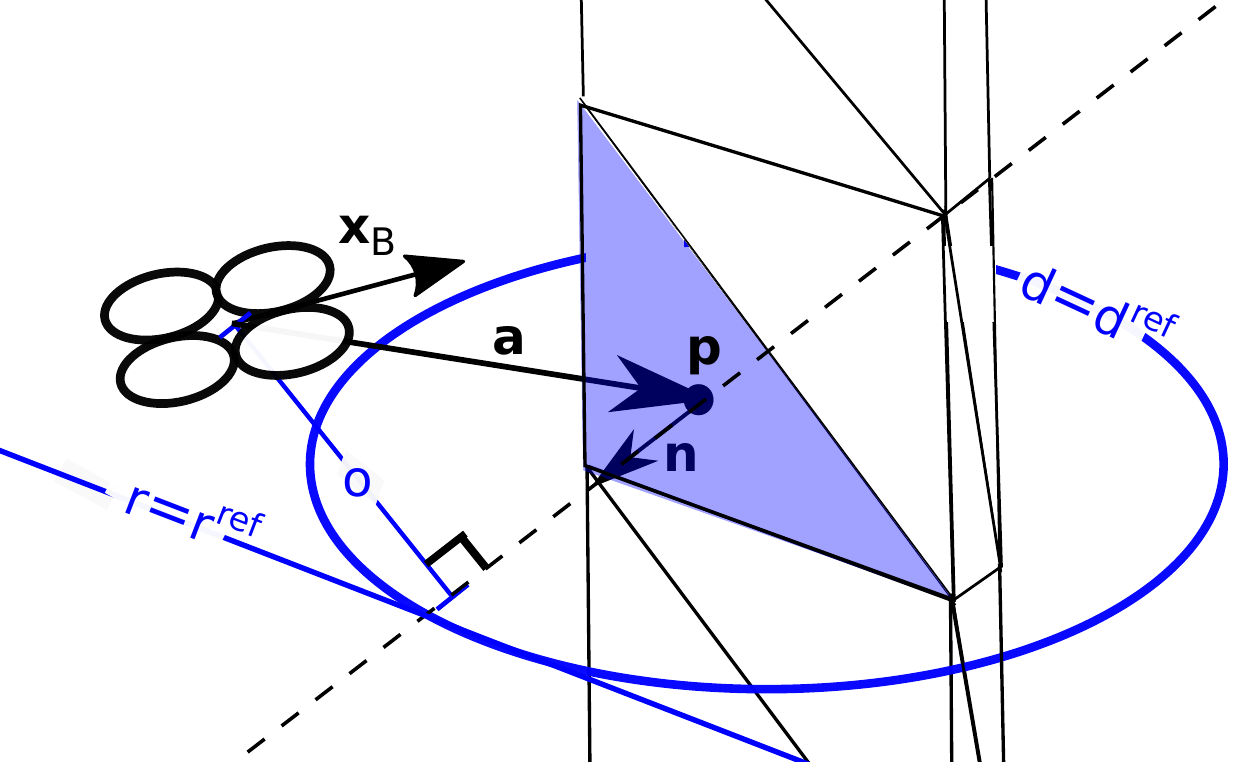}
    \caption{Graphical depiction of the visual costs for the VT-NMPC for inspecting a surface of normal $\mathbf{n}$. The intersection between the line $r=r^\textrm{ref}$ , the circle $d=d^\textrm{ref}$ and the normal vector $\mathbf{n}$ represent the optimal position that the drone should attain during inspection relative to the surface. }
    \label{fig:methodoverview}  
\end{figure}

The least-square objective of the proposed \ac{VT-NMPC} with nonlinear costs and constraints can be formulated in discrete time as follows:

\begin{subequations} \label{eq:NMPC}
\begin{align}
	\minimize_{\mathbf{x}_k,\mathbf{u}_k} \; & \frac{1}{2} \Biggl\{\sum_{k = 0}^{N_c-1} \left( \mathcal{C}_{h}
	+\mathcal{C}_{{d}} +\mathcal{C}_{{r}} + \mathcal{C}_{{o}} + \mathcal{C}_{\mathbf{x}^*}+ \mathcal{C}_{\mathbf{u}}\right) + \mathcal{C}_{N_c} 
	\Biggr\} \label{eq:NMPC1} \\
	\textrm{s.t.} \; & \mathbf{x}_{k+1} = \mathtt{f_d}(\mathbf{x}_k,\mathbf{u}_k), \hspace{0.5cm} k \in [0, N_c-1],  \label{eq:NMPC2} \\
    & \mathbf{u}_{k,\textrm{min}} \leq \mathbf{u}_k \leq \mathbf{u}_{k,\textrm{max}}, k \in [0, N_c-1], \label{eq:NMPC4}
\end{align}
\end{subequations}
where the terms $\mathbf{u}_{k,\textrm{min}} \leq \mathbf{u}_{k,\textrm{max}} \in \mathbb{R}^{4}$ in \eqref{eq:NMPC4}, specify the lower and upper bounds on the controls, respectively. The terms inside the summation in \eqref{eq:NMPC1} together account for the stage cost. Amongst them, the first four costs, namely $\mathcal{C}_{h}$, $\mathcal{C}_{d}$, $\mathcal{C}_{r}$, $\mathcal{C}_{o}$ represent the cost associated with visual tracking objective and are expressed in the following form: 
\begin{align} 
	\mathcal{C}_{h} &= (h_k - h^{\textrm{ref}})^T\mathrm{w}_{h}(h_k - h^\textrm{ref}), \label{eq:NMPC1_stageCostVisual_head} \\
	\mathcal{C}_{d} &= (d_k - d_k^{\textrm{ref}})^T\mathrm{w}_{d}(d_k - d^{\textrm{ref}}),  \label{eq:NMPC1_stageCostVisual_dist} 
	\\
	\mathcal{C}_{r} &= (r_k - r_k^{\textrm{ref}})^T\mathrm{w}_{r}(r_k - r^{\textrm{ref}}),  \label{eq:NMPC1_stageCostVisual_roi}
	\\
	\mathcal{C}_{o} &= (o_k - o_k^{\textrm{ref}})^T\mathrm{w}_{o}(o_k - o^{\textrm{ref}}),  \label{eq:NMPC1_stageCostVisual_oi}
\end{align}
The first visual tracking cost, $\mathcal{C}_{h}$ essentially controls the heading of the drone, which penalizes the deviation of the inspection area's centroid from the drone's \ac{FOV}-center. Accordingly, a heading function $h$ is formulated as a dot product between a vector representing the forward direction of the drone in the Earth-fixed frame and a vector $\mathbf{a}$ that connects the drone to the center of the inspection area in the X-Y plane, as illustrated in Fig.~\ref{fig:methodoverview}. Consequently, the heading function is expressed as follows:
\begin{align}\label{eq:heading_function}
    h &= (R_{EB} \; \mathbf{x}_B)\cdot \mathbf{a}_{xy},
\end{align}
where the vector $\mathbf{x}_B = [1, 0, 0]^T$, and $\mathbf{a}_{xy} = [a_{x}, a_{y}, 0]^T$ reflects a vector comprising the $x-y$ components of the original vector $\mathbf{a}$, which is given as:
\begin{equation}\label{eq:a}
    \mathbf{a} = \begin{bmatrix}
           a_{x} \\
           a_{y} \\
           a_{z}  
         \end{bmatrix}
    = \begin{bmatrix}
           p_{x}-x \\
           p_{y}-y \\
           p_{z}-z  
         \end{bmatrix}.
\end{equation}
where $\mathbf{p} = [p_x, p_y, p_z]^T$ refers to the center of the inspection area. Moreover, the reference value for the heading function $h^\textrm{ref}$ is set to $1$ so that the two vectors are aligned.  

The second visual tracking cost, $\mathcal{C}_{d}$ controls the position of the drone such that the drone stays within a specified radius from the center point $\mathbf{p}$. This cost penalizes the deviation of the current Euclidean distance ($d$) from its specified reference value ($d^\textrm{ref}$). Also, the Euclidean distance is calculated in the X-Y plane as follows:
\begin{align}\label{eq:euc_dist}
    d &= \| \mathbf{a}_{xy}\|.
\end{align}
The reference Euclidean distance $d^\textrm{ref}$ is the distance that the user desires to maintain from the wind turbine for safety. In this work, $d^{\textrm{ref}}$ is set to $7$m, which has been the de-facto reference distance in the literature for wind turbine inspection \cite{lidar2,stokkeland,ICUAS2020}. 
The third visual tracking cost, $\mathcal{C}_{r}$ ensures that the drone stays within the desired region of interest, which is maintaining a specific distance from the inspection surface along its surface-normal in the X-Y plane. The region of interest function $r$ is calculated by taking the dot product of the vector $\mathbf{a}_{xy}$ and the normal vector to the surface being inspected:
\begin{align}\label{eq:roi_functon}
    r &= \mathbf{a}_{xy}\cdot \mathbf{n},
\end{align}
where the normal vector is given by $\mathbf{n} = [n_x, n_y, 0]^T$. The reference distance to be maintained ($r^\textrm{ref}$) is set equal to $d^{\textrm{ref}}$, i.e., $r^{\textrm{ref}}$ = $d{^\textrm{ref}}$ = 7m. It is to be noted that, the region of interest function $r$ together with the distance function $d$ guide the drone to the desired position -- at a specified distance from the inspection surface centroid --, while aligning with the positive direction of the surface-normal. \\
The final visual tracking cost, $\mathcal{C}_{o}$ comprises of an orthogonality function $o$ (visually depicted in Fig.~\ref{fig:methodoverview}), which is defined as the magnitude of the orthogonal projection of the vector $\mathbf{a}$ on the surface normal $\mathbf{n}$, and is expressed as follows:
\begin{align}\label{eq:orth_functon}
    o &= \|(\mathbf{a}-(\mathbf{a}\cdot \mathbf{n} )\mathbf{n})\|,
\end{align}
in essence, the role of this cost is to align the drone with the surface normal, hence, $o^{\textrm{ref}}$ is set to $0$. The last two stage costs in \eqref{eq:NMPC1} are the costs associated with states and control, penalizing the deviations of the predicted states and control trajectories from their references. They are given as follows:
\begin{align} \label{eq:NMPC1_stageCostStateControl}
	\mathcal{C}_{\mathbf{x}^{\textrm{*}}} &= (\mathbf{x}_k^{\textrm{*}} - \mathbf{x}_k^{\textrm{* ref}})^T\mathbf{W}_{\mathbf{x}^{\textrm{*}}}(\mathbf{x}_k^{\textrm{*}} - \mathbf{x}_k^{\textrm{* ref}}), \\
	\mathcal{C}_{\mathbf{u}} &= (\mathbf{u}_k - \mathbf{u}_k^{\textrm{ref}})^T\mathbf{W}_{\mathbf{u}}(\mathbf{u}_k - \mathbf{u}_k^{\textrm{ref}}),
\end{align}
where $\mathbf{x}^{\textrm{*}} = [u,v,w, q_x,q_y]^T \in \mathbf{x}$. The following trajectories are selected for a smooth response from the drone:
\begin{align} 
	\mathbf{x}^{\textrm{* ref}} &= \mathbf{x}_{N_c}^{\textrm{* ref}} =  \left[u_r,v_r,w_r,0,0 \right]^T, \label{eq:ref_x_NMPC} \\ 
	\mathbf{u}^{\textrm{ref}} &=  \left[0,0,0, mg \right]^T. \label{eq:ref_u_NMPC}
\end{align}
The weights defined in the stage cost are given by $ \mathrm{w}_{h} \in \mathbb{R}^{1 \times 1} $, $\mathrm{w}_{d} \in \mathbb{R}^{1 \times 1}$, and $  \mathrm{w}_{r} \in \mathbb{R}^{1 \times 1} $, $ \mathbf{W}_{\mathbf{x}^{\textrm{*}}} \in \mathbb{R}^{5 \times 5} $, $  \mathbf{W}_{\mathbf{u}} \in \mathbb{R}^{4 \times 4}$. The weight matrices associated with state and control costs are selected to be diagonal and positive-(semi) definite matrices. 
Finally, the second component in \eqref{eq:NMPC1} refers to the terminal cost which is given by: 
\begin{align} \label{eq:NMPC1_terminalCost}
	& \mathcal{C}_{\mathbf{N_c}} = (\mathbf{x}_{N_c}^{\textrm{*}} - \mathbf{x}_{N_c}^{\textrm{* ref}})^T\mathbf{W}_{N_c}(\mathbf{x}_{N_c}^{\textrm{*}} - \mathbf{x}_{N_c}^{\textrm{* ref}}).
\end{align}
The above cost penalizes the finite nature of the prediction horizon, which caters to the stability of the overall optimization problem. Herein, $ \mathbf{W}_{N_c} \in \mathbb{R}^ {5 \times 5} $ is the corresponding positive-(semi) definite weight matrix which is also selected as diagonal. 
In terms of the implementation of the \ac{VT-NMPC} method, the weight matrices are selected based on the method proposed in \cite{deepmodel_mohit}. The incorporated active exploration approach renders automated tuning of the controller weights based on the drone’s performance during tuning trials. The method essentially utilizes an intelligent trial-and-error-based strategy that utilizes retrospective knowledge to improve the next set of controller weight attempts, balancing exploration and exploitation in an automated fashion. Accordingly, the following weights are achieved:
\begin{align*}
    \mathrm{w}_h &= 80, \; \mathrm{w}_d = 30, \mathrm{w}_r = 25, \mathrm{w}_o = 60\\ 
	\mathbf{W}_{\mathbf{x}^{\textrm{*}}} &= \textrm{diag}(0.3,0.3,1,80,80) \\
	\mathbf{W}_{\mathbf{u}} &= \textrm{diag}(1,1,0.25,0.03), \mathbf{W}_{N_c} =1.5\mathbf{W}_{\mathbf{x}}.
\end{align*}
Additionally, the following constraints are imposed on the control input to obtain a smooth response,
\begin{align} 
    -100\ (^{\circ}/\textrm{s}) \leq & \; p, q, r \leq 100\ (^{\circ}/\textrm{s}), \label{eq:const_rates} \\
    0.3mg\ \text{(N)} \leq & \quad T \quad \leq 2mg\ \text{(N)}. \label{eq:constT}
\end{align}
%

\subsection{Implementation Details}
The optimization problem underlying the \ac{VT-NMPC} is solved by utilizing the direct multiple shooting method with a shooting grid size of $0.01$s, to enable real-time feasibility. Direct solution methodology is adopted over the indirect methods, as the former can account for both inequality and equality constraints. Subsequently, the resulting discretized optimal control problem (OCP) reduces to a sequential quadratic program which is further solved via the generalized Gauss-Newton method with the help of a special real-time iteration (RTI) scheme proposed in \cite{DIEHL2002577}. Moreover, ACADO toolkit \cite{acado}, along with qpOASES solver 
is utilized as the solution platform for solving the OCP in \eqref{eq:NMPC}
as it incorporates the direct multiple shooting method and the RTI approach altogether. Additionally, the prediction horizon $N_c = 30$ and a sampling time of $0.01$s for computational tractability.

\section{Simulations}

\label{sec:results}

\begin{figure}
\centering
\includegraphics[width=.46\textwidth]{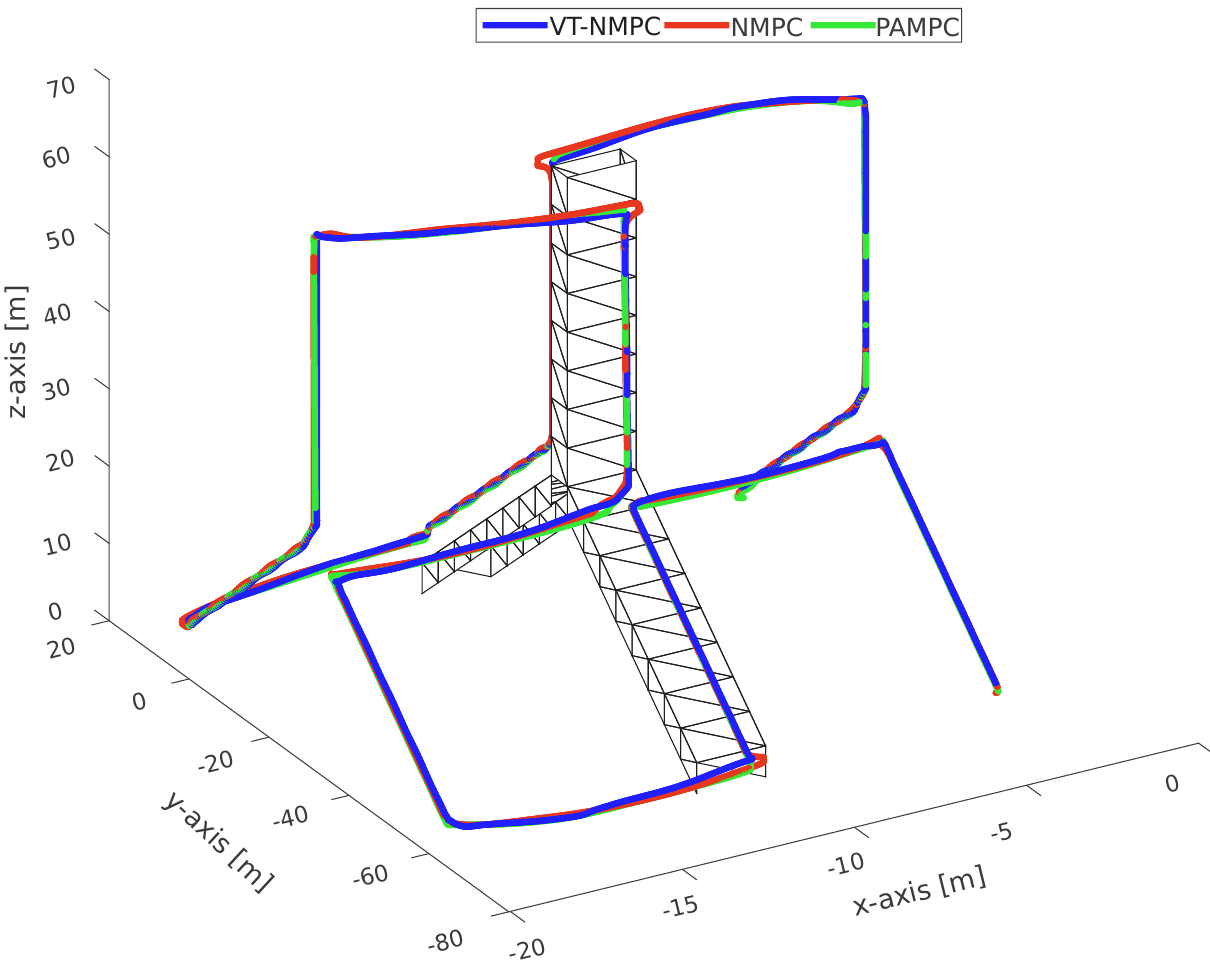}
\caption{Obtained optimal 3-D trajectory of the drone during inspection using the proposed novel VT-NMPC, traditional NMPC and PAMPC.}\label{3d}
\end{figure}

This section demonstrates the results of a wind turbine inspection scenario conducted in a Gazebo simulation environment. A sinusoidal wind disturbance with a time period of $10$s and a standard deviation of $0.5$m/s is applied in the y-direction to further model a wind farm environment. The results are obtained for three different mean wind speed levels, namely, $0$m/s, $4$m/s, and $7$m/s. Note that only the plots for $4$m/s mean wind speed are presented for brevity. 
The efficacy of the proposed VT-NMPC method is compared with the PAMPC \cite{falanga2018pampc} and the conventional NMPC. Both the PAMPC and NMPC follow a reference trajectory consisting of 3-D positions. Besides, the NMPC requires the reference yaw angle, while the PAMPC uses the inspection point as a reference. The reference position trajectory is obtained as an offset from the inspection point, at a distance equal to $d^{\mathrm{ref}}$ in the direction of the surface normal, while the reference yaw angle is precomputed based on the surfaces' normal $\mathbf{n}$. The performance of all three methods is evaluated in terms of three metrics, i.e., safety, coverage, and inspection quality. The obtained results are summarized in Table~\ref{table:sim}.  Safety implies the distance from the inspection surface that the drone maintains. Coverage illustrates the ability to view all the triangular mesh elements that underly the turbine model. Inspection quality refers to the quality of the obtained images. The obtained 3-D trajectory of the drone using different methods is shown in Fig. \ref{3d}. Note that the drone's average speed during the inspection is around $0.7$m/s.

%


\subsection{Safety Metric}

First, a safety region requirement is defined as a boundary from the desired reference, $d^{ref}$, in the positive and negative direction. A 1-meter range (safety margin) is selected, beyond which the drone is considered to be violating this requirement. Accordingly, the safety metric is calculated as follows:

\begin{equation}
    \text{safety metric (SM)} = 
    \begin{cases}
      100\%, &   |d - d^{ref}| < \text{safety margin}\\
      0, & \text{otherwise}
    \end{cases}
    \label{eq:safety}
\end{equation}
which is then averaged over the entire trajectory. 
 The VT-NMPC, PAMPC, and NMPC manage to maintain comparable distances on average with $7.03$m, $7.07$m, and $7.06$m respectively. Nevertheless, the amplitude of fluctuations for the VT-NMPC is significantly lower than the NMPC and PAMPC, obtaining a minimum distance of $6.15$m compared to $5.84$m for the PAMPC and $5.63$m for the NMPC.

 The ability of the VT-NMPC to maintain the desired distance compared to the other two methods can be attributed to the inclusion of distance cost in \eqref{eq:NMPC1_stageCostVisual_dist}.  In essence, it explicitly penalizes the distance from the inspection surface compared to the other two which are tracking-error driven, i.e., separately minimizes the error in $x$, $y$, and $z$ components.

\subsection{Coverage}
During the inspection, it is required that all surface areas of the parts being inspected are to be captured by the drone's camera. We use the triangular mesh of the generated wind turbine model to measure coverage by checking whether each triangle that comprises the model has been viewed. 
For each time instance along the trajectory, we check the triangles that are within \ac{FOV} of the drone by projecting its vertices onto the drone's image plane using a pinhole camera projection. 
Subsequently, we calculate the coverage as the ratio between the total number of viewed triangles to the total number of triangles.
A $100\%$ coverage is achieved by all the three methods with $0$m/s wind condition. When introducing a wind speed of $4$m/s, both the VT-NMPC and PAMPC maintain a $99.8\%$ coverage value, while the NMPC's coverage reduces to $91.6\%$. 
The presence of the visual feedback in the vision-based control methods (VT-NMPC and PAMPC) allows the drone to maintain high coverage for all the three wind levels. 
 
\begin{table}
    \centering
    \begin{tabular}{|C{2.0cm}|C{1.5cm}|C{1.5cm}|C{1.5cm}|C{1.5cm}|C{5.5cm}| }
        \cline{2-4}
        \multicolumn{1}{c|}{} & \textbf{NMPC} & \textbf{PAMPC} & \textbf{VT-NMPC} \\
\hline
        Coverage &  $91.6\%$ & $ 99.8\%$ & $\textbf{100\%}$ \\
        \cline{2-4}
        \hline
        
        \multirow{1}{*}{Safety (SM)}  
          & $75.0\%$ &$84.7\%$ &  $\textbf{100\%}$ \\
        \hline        
         \multirow{1}{*}{Quality (CM)}     & $98.5\%$ & $\textbf{99.9\%}$ & 
        $\textbf{99.9\%}$ \\
        \hline
    \end{tabular}
    \caption{Simulation results for the defined performance indices. Tests were conducted at a wind speed of 4 m/s.}
    \label{table:sim}
\end{table}

\subsection{Inspection Quality }

An additional requirement is the quality of the images collected throughout the inspection operation. The wind turbine blade should be always centered within the image frame.
This criterion is evaluated by the centering metric (CM), which is calculated as $\mathrm{CM} = h$, where $h$ is the heading function defined in \eqref{eq:heading_function}. As can be seen in Table~\ref{table:sim}, high average values for CM are achieved for vision-based control methods, i.e., $99.9\%$ for the VT-NMPC and the PAMPC.

\section{Real-World Experiments} \label {sec:results_r}


\subsection{Experimental Setup}

The real-world experiments are conducted in a motion capture system environment having 16 cameras. It is important to note that the use of this system is consistent across all methods and should not have affected the presented comparative study. Additionally, a blowing tunnel is incorporated to provide disturbances with a wind speed of $4$ (m/s), reflecting the operational conditions during wind turbine inspections.\textbf{}
As such, we develop in-house a small-scale quadrotor, having a total mass of $1090$g and a rotor-to-rotor diagonal dimension of $250$ mm. The drone is equipped with an Nvidia Jetson TX2 onboard computer that runs the trajectory generation and VT-NMPC codes online. Also, a Pixhawk flight controller running the PX4 firmware renders the controller mapping from thrust and body rates to rotor speeds. In terms of the experimental setup, the drone is tasked with inspecting a single blade of the turbine inspected in Section VI. Besides, the blade is scaled by a factor of $1/10$, $1/10$, and $1/15$ in x, y, and z dimensions, respectively, and $d^{\mathrm{ref}}=0.5$m is set, to cater to the limited testing space. 


\subsection{Experimental Results}

Next, we present the experimental results, wherein the proposed VT-NMPC method and the baselines are evaluated in a similar setting to the simulation study. As such, the tracked trajectories from the three methods are utilized to compute the defined metrics, i.e., safety, coverage, and inspection quality.


\begin{figure}
\centering
\includegraphics[width=.45\textwidth]{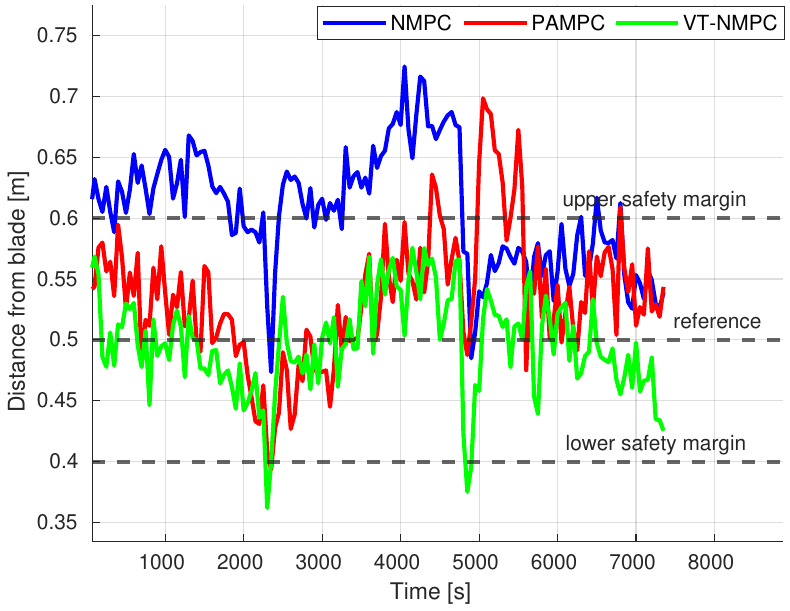}
\caption{A plot of the distance from the blade throughout the trajectory, showing the upper and lower safety margins and the reference ($d^{ref}$). }  \label{fig:distance_r}
\end{figure}

\begin{figure}
\centering
\includegraphics[width=.45\textwidth]{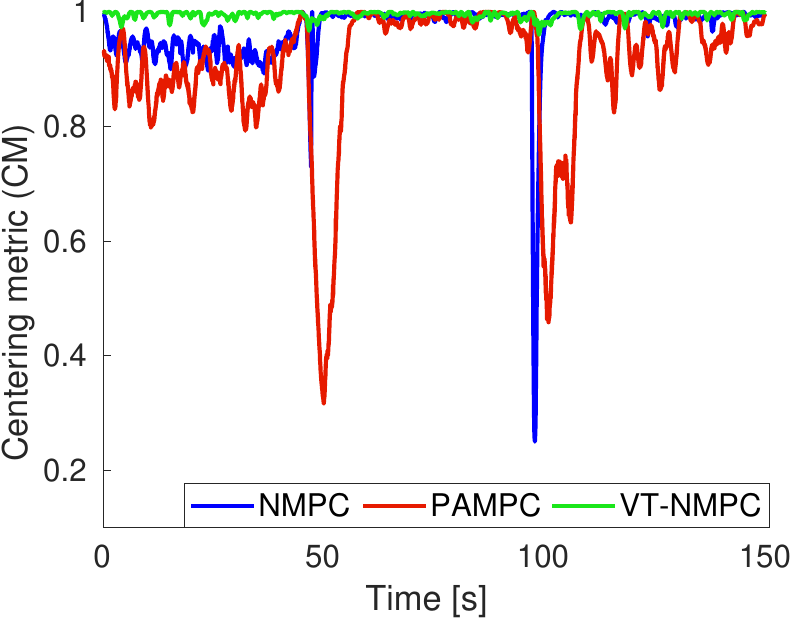}
\caption{A plot of the centering metric (CM) vs. time. VT-NMPC consistently tracks the center of the inspected surface area, even under wind disturbance, as opposed to PAMPC and NMPC.}  \label{fig:cm_r}
\end{figure}

For the specific experimental run, the obtained coverage values from the VT-NMPC, PAMPC, and conventional NMPC are $100\%$, $90\%$, and $96\%$, respectively. This validates the superiority of the added visual tracking costs within the VT-NMPC method. Then, to illustrate the safety metric, the resulting absolute distance error throughout the trajectory is shown in Fig.~{\ref{fig:distance_r}}. Note that a safety margin of 10cm is selected in real-world experiments due to the limited testing area. As can be seen, the proposed VT-NMPC method on average maintains the lowest error to the reference in comparison to the baselines. Accordingly, it enables the drone to stay within the safety limits for 98\% of the run time, even for such a small safety margin. It may be worth noting that for a marginal increase of 5cm in the safety margin, an SM = 100\% is realized for the VT-NMPC method, while the other competitors still violate the limits for substantial run time. The centering metric performance can be observed in Fig.~{\ref{fig:cm_r}}. As visualized, the real-world results manifest a dominating performance of VT-NMPC compared to the PAMPC and conventional NMPC. One may note a performance drop for the PAMPC method in the real-world results. This is attributed to the conservative tuning of the perception cost, which otherwise resulted in instability. The main reason for this behavior is the complexity of the perception cost, which also has the possibility of a division by zero during the projection step. Furthermore, the computational effort is measured for the three controllers. The execution time is lowest for the conventional NMPC (5 milliseconds), followed by VT-NMPC (10 milliseconds), then PAMPC (13 milliseconds), implying that the benefits of vision-based MPC methods come at an additional computational burden. Nevertheless, the proposed VT-NMPC method does show a 30\% performance improvement over the baseline PAMPC, which can be attributed to the simplicity of its visual costs.

\begin{table}
\centering
    \label{tab:compare}
    \begin{tabular}{|C{2.0cm}|C{1.5cm}|C{1.5cm}|C{1.6cm}| }
    
        \cline{2-4}
        \multicolumn{1}{c|}{} & \textbf{NMPC} & \textbf{PAMPC} & \textbf{VT-NMPC} \\
        \cline{1-4}
        \hline
        Coverage   & $98.0\% $ & $91.0\%$ & $\textbf{99.8\%}$ \\
        \hline
        \multirow{1}{*}{Safety (SM)}  &  $43.0\%$ & $89.9\%$ &$\textbf{98.0\%}$\\
        \hline
        \multirow{1}{*}{Quality (CM)}  & $97.0\% $ & $90.8\%$ & $\textbf{99.5\%} $ \\
        \hline
    \end{tabular}
    \caption{Statistical analysis of the experimental results. For each metric, the average values in percentage are calculated based on three experimental runs. $d^\textrm{ref}$ is specified as $0.5$m and the safety margin is 0.1 m.
    }
    \label{table:stats}
\end{table}

\section{Conclusions}
\label{sec:conclusion}



In this work, an automated wind turbine inspection framework incorporating a novel NMPC method with visual tracking objectives is presented. Multiple simulations and real-world tests validated that the proposed VT-NMPC method outperforms traditional MPC methods in several aspects. When compared to the PAMPC, the heading-control term is more intuitive, facilitating easier implementation and shows better performance in achieving full coverage and high inspection quality. Moreover, the VT-NMPC method shows promising improvements in obtaining better image quality by maintaining an optimal relative pose to the inspection surface. Besides, the proposed method infers a potential benefit of directly incorporating reference surface's pose in the MPC formulation rather than utilizing a position trajectory for the drone during inspection. It is worth nothing that even though the proposed method is exclusive to wind turbine inspection, it can be extended to a broader inspection spectrum by replacing the wind turbine-specific path planner.
As part of future work, we intend to collaborate with a wind turbine manufacturer to conduct outdoor testing of the proposed framework.

\section{Acknowledgement}
{
This work is supported by EIVA a/s and Innovation Fund Denmark under grants 2040-00032B and 2035-00052B. The authors would like to thank Peter Harling Lykke for his support during the real-world experiments. Furthermore, the authors would further like to acknowledge Upteko Aps for bringing the use-case challenge.
}

\bibliography{refs.bib}

\addtolength{\textheight}{-12cm}   





\end{document}